%% file: cvpr.tex
\documentclass[10pt,twocolumn,letterpaper]{article}

\usepackage{cvpr}              

\usepackage{graphicx}
\usepackage{amsmath}
\usepackage{amssymb}
\usepackage{booktabs}

\usepackage[pagebackref,breaklinks,colorlinks]{hyperref}
\usepackage[inline]{enumitem}
\usepackage{amssymb}
\usepackage{algpseudocode}
\usepackage{algorithm}
\usepackage{graphicx}
\usepackage{float}
\usepackage{amsmath}
\usepackage{mathrsfs}
\usepackage{multirow}
\usepackage{xcolor}

\definecolor{mygreen}{RGB}{103,171,159}
\definecolor{myorange}{RGB}{240,144,66}

\newcommand{\abbrModel}{GSE-GCN}

\usepackage[capitalize]{cleveref}
\crefname{section}{Sec.}{Secs.}
\Crefname{section}{Section}{Sections}
\Crefname{table}{Table}{Tables}
\crefname{table}{Tab.}{Tabs.}

\def\cvprPaperID{7217} 
\def\confName{CVPR}
\def\confYear{2022}

\begin{document}

\title{IntentVizor: Towards Generic Query Guided Interactive Video Summarization}

\author{
Guande Wu \thanks{equal contribution} \qquad Jianzhe Lin \footnotemark[1] \qquad Claudio T. Silva\\
New York University\\
\{guandewu, jianzhelin, csilva\}@nyu.edu

}
\maketitle

\begin{abstract}
	\input sections/abstract
\end{abstract}


\input{sections/introduction}

\input{sections/related}
\input{sections/model}
\input{sections/experiments}
\input{sections/conclusion}
\input{sections/acknowledgement}
{\small
\bibliographystyle{ieee_fullname}
\bibliography{main}
}

\end{document}

%% file: sections/abstract.tex
The target of automatic video summarization is to create a short skim of the original long video while preserving the major content/events. There is a growing interest in the integration of user queries into video summarization or query-driven video summarization. This video summarization method predicts a concise synopsis of the original video based on the user query, which is commonly represented by the input text. However, two inherent problems exist in this query-driven way. First, the text query might not be enough to describe the exact and diverse needs of the user. Second, the user cannot edit once the summaries are produced, while we assume the needs of the user should be subtle and need to be adjusted interactively. To solve these two problems, we propose IntentVizor, an interactive video summarization framework guided by generic multi-modality queries. The input query that describes the user's needs are not limited to text but also the video snippets. We further represent these multi-modality finer-grained queries as user `intent', which is interpretable, interactable, editable, and can better quantify the user's needs. In this paper, we use a set of the proposed intents to represent the user query and design a new interactive visual analytic interface. Users can interactively control and adjust these mixed-initiative intents to obtain a more satisfying summary through the interface. Also, to improve the summarization quality via video understanding, a novel Granularity-Scalable Ego-Graph Convolutional Networks (GSE-GCN) is proposed. We conduct our experiments on two benchmark datasets. Comparisons with the state-of-the-art methods verify the effectiveness of the proposed framework. Code and dataset are available at \url{https://github.com/jnzs1836/intent-vizor}.

%% file: sections/introduction.tex
\section{Introduction}
\begin{figure}[t]
\centering
\includegraphics[width=0.45\textwidth]{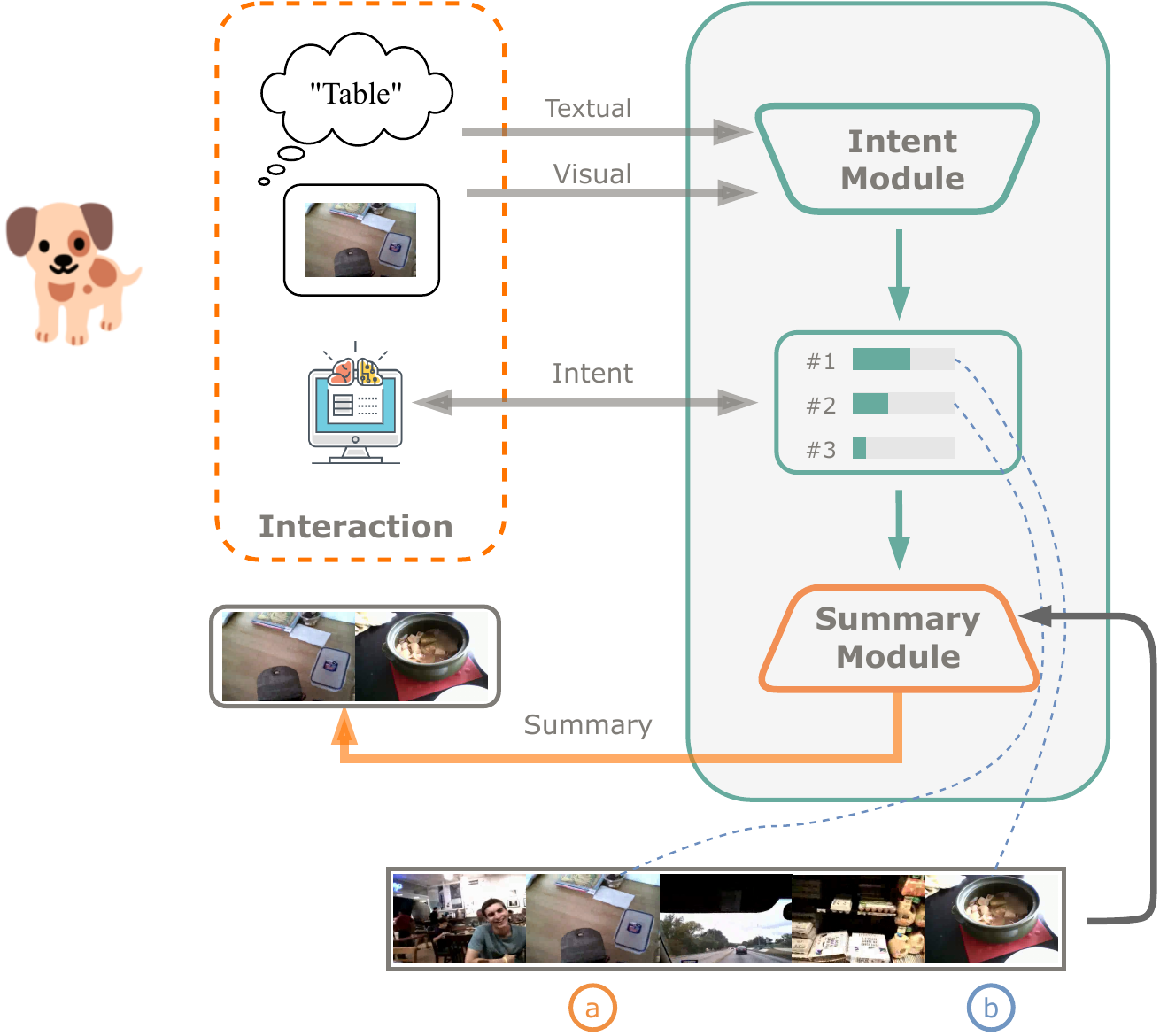} 
\caption{Illustration of our IntentVizor framework. 
We take query ``Table" as an example. Generic queries, including text/video snippets related to ``Table" are inputs of the model. The intent module transforms these queries into a probability distribution over the basis intents, followed by the summary module, which generates a video summary by combing the basis intents and their probability values. As the user can find the underlying visual semantic meaning of each basis intents (e.g., in the figure, basis intent \#1: the dining table; \#2: the working table), they can adjust the distribution of these basis intents through our proposed interface (Fig. \ref{fig:prototype-overview}) to satisfy their needs, and the final generated summaries can be updated accordingly/iteratively.
}

\label{fig:intro}
\end{figure}

With the online explosive video content, an increasing need has been identified for automatic video summarization in recent years.  
Traditional video summarization methods usually generate concise/representative summary that contains the entities and events with high priority from the video and with low repetition and redundancy using unsupervised \cite{jung2019discriminative, mahasseni2017unsupervised,chu2015video,panda2017diversity,yaliniz2019unsupervised,zhang2016context,kanafani2021unsupervised}, supervised \cite{papalampidi2020movie, zhu2020dsnet,zhang2016video,zhao2018hsa,fajtl2018summarizing,park2020sumgraph,zhao2019property} and reinforcement learning ways\cite{apostolidis2020ac, zhou2018deep}. However, such a summary cannot satisfy the needs of users and be of low practical value. As the elongated video, especially when captured in the realistic scenario, may cover a wide range of topics, only fractional content of specific topics will meet the user's needs. Based on this observation, the user query-driven summarization model, which considers the user's preference, has gradually attracted researchers' attention.



The basic idea for query-driven summarization is to use the text query to guide the generation of video summaries. A popular dataset for this query-driven summarization was the textual query dataset, proposed by Sharghi et al. \cite{sharghi_query-focused_2017}.
The summarization model proposed in the paper was trained to predict a subset of the video shots (5 seconds per shot) closely related to the textual query. 
For the follow-up works, the attention mechanism \cite{xiao_convolutional_2020, xiao_query-biased_2020, messaoud_deepqamvs_2021} and generative adversarial networks\cite{zhang_query-conditioned_2018} based summarization models are also introduced to achieve better summarization performance. However, the performance of these models was still not satisfying as the textual query is not enough to represent the users' preferences. To be more specific, first, the user cannot express their detailed needs with few fixed input textual queries at the very beginning of summarization. They may have multiple needs and want to adjust the priority of different needs. Second, the textual query can be ambiguous. People can have different understandings of a word in the communication, let alone the model trained on a fixed word dictionary. Therefore, the model should be {interactive} to loop users into the summarization, and other query formats (e.g., visual query) should be considered to better represent the user preference with lower ambiguity.

To propose a generic model for queries from different modalities and allow users to interact during the summarization process, in this paper, we propose a novel framework named as IntentVizor. We borrow the concept \textbf{\textit{Intent}} from the Information Retrieval (IR) community to define the users' need, independent of the query modalities\cite{kofler2016user, zhang2019generic, broder2002taxonomy}. However, our {{intent}} differs from the traditional definition in IR with different representation and extraction: 
    (1) We represent the {{intent}} by an adjustable distribution over the basis intents rather than the pre-defined categories\cite{broder2002taxonomy} , taxonomies\cite{yin2010building, broder2007robust} or in a distributed representation space\cite{zamani2016estimating,guo2016deep}; The basis intents are defined as the learned and basic components of the user's needs. Compared with the traditional definitions\cite{broder2002taxonomy,yin2010building, broder2007robust,zamani2016estimating,guo2016deep}, our method enables interactive manipulation, satisfying the user's diverse and subtle needs. 
    (2) We extract a unified {{intent}} from the queries of different modalities instead of only the textual query to avoid the ambiguity problem as mentioned before. 

The intentVizor framework consists of two modules, i.e., the intent module for extracting the intent from the query and the summary module for summarizing the video with the intent.
To effectively correlate the video features with the generic query/intent in the two modules, we design a flexible network structure named Granularity-Scalable Ego-Graph Convolutional Network (GSE-GCN). This GSE-GCN will work as a shared backbone for both the summary module and the intent module. Besides this backbone, the two modules each has an intent head and a summary head separately.

To sum up, we structure our contributions as follows:
\begin{itemize}
    \item To the best of our knowledge, our IntentVizor framework is the first attempt to introduce generic queries to better satisfy the user's diverse needs. We also propose a novel dataset for the visual-query-guided video summarization based on UTE videos.
    \item We formulate the video summarization as an interactive process, where the user can fine-tune its intent iteratively with our proposed novel interface.

    \item We propose a novel \abbrModel \space structure to effectively correlate the generic queries of multi-modalities with the input video.
\end{itemize}

%% file: sections/related.tex
\section{Related Works}

\subsection{Summarizing Video with User Intent}
Integrating the user query in video summarization has long been a hot topic. Previous approaches often represent the query by the textual concepts. Sharghi et al. built the first query-focused video summarization dataset based on the UTE video dataset and proposed an evaluation protocol based on the shot-level semantic tags, which became the standard protocol for the research community \cite{sharghi_query-focused_2017}. Zhang et al. proposed a generative adversarial network (GAN) to address the problem \cite{zhang_query-conditioned_2018}. Xiao et al. defined the task as a shot-query similarity problem and proposed a convolution-based network to capture the local and global information\cite{xiao_convolutional_2020}. We also calculate the distance-based similarity between the video shots and user intent.
Jiang et al. employed a multilevel self-attention module and a variational autoencoder (VAE) to add user-oriented diversity and stochastic factors\cite{jiang_hierarchical_2019}. 
Although their work also paid attention to the user intent, they did not allow the interactive adjustment of the intent. Instead, recent works gradually noticed the importance of the user feedback and user interaction \cite{jin2017elasticplay, del2017active}. 
However, these works still lack a flexible method for the user to control the interaction process. To solve this problem, we achieved this interaction through intent (the controllable variable for user) adjusting in this paper.

\subsection{Graph Convolutional Networks for Videos}
Graph convolutional networks have been widely applied on various video analysis tasks. The approaches can be roughly divided into two categories based on the graph type, i.e., spatio-temporal graph and snippet graph. Spatio-Temporal graph views the video as a graph of region proposals or objects in the spatio-temporal domain. 
Liu et al. represented the video as a space-time region graph and applied a GCN to perform the action recognition \cite{wang2018videos}. 
Yan et al. modeled the dynamic body joints as a spatio-temporal graph to estimate the human pose \cite{yan2018spatial}. Similar approaches were also employed in a wide range of tasks including, action recognition\cite{mavroudi2019neural, wang2021weakly, ghosh2020stacked}, human re-identification\cite{liu2021spatial}, gaze prediction\cite{fan2019understanding}, and video captioning\cite{pan2020spatio}. Unlike the above works, the snippet graph correlated the snippets (`snippets' are `segments' in our paper, as mentioned before) by their semantic and temporal relationships. 
Zeng et al. built a graph of the temporal 1-D proposals to perform the temporal action localization\cite{zeng2019graph}.
Xu et al. constructed a snippet graph and designed an efficient edge convolution method to detect the temporal action\cite{xu2020g}. We borrowed their edge convolution operation when we introduced a hybrid graph with the user intent to align the video segments and user intent.


%% file: sections/model.tex
\section{IntentVizor Framework}
Our IntentVizor framework targets at (1) interactive control over the video summarization process; (2) support of the generic multi-modality query. 
This section first shows that the two requirements can be satisfied by modelling the multi-modality queries as a unified and interactive user intent. Then, we will describe \abbrModel, which is designed to better deal with multi-modality queries.


\begin{figure*}[t]
\centering
\includegraphics[trim=0 0 0 0,width=0.95\textwidth]{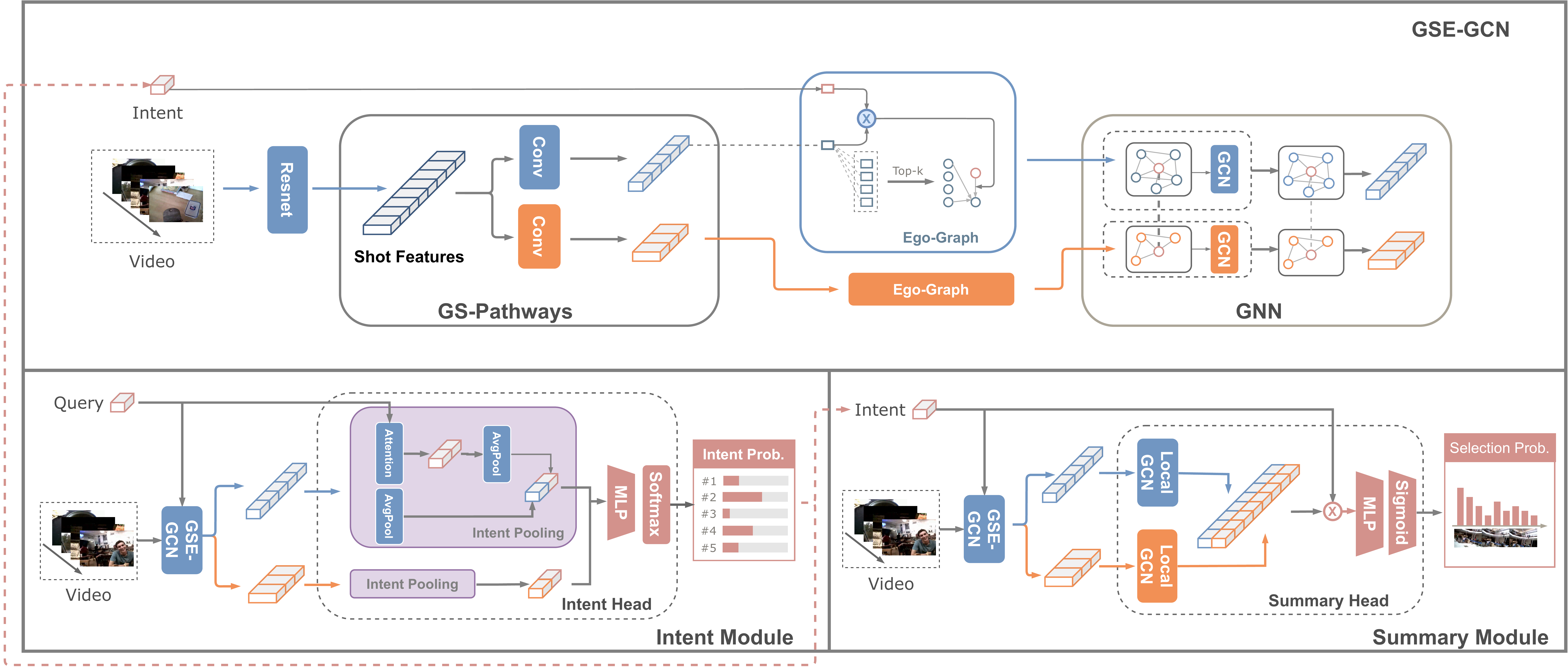} 

\caption{\textbf{GSE-GCN} exploits two notions i.e., GS-Pathway and Ego-Graph. The input video will be processed by two convolutional networks to produce two segment-level feature sequence of coarse and fine granularity. Then, each sequence will be processed to generate a Ego-Graph, where the intent/query vertex is an ego-vertex with all the video segments are connected. After feeding the graph into GCN, the two pathways will be produce the corresponding segment-level features. \textbf{Intent Head} pools the segment features into a distributed representation, which will be processed by a MLP with softmax to produce the intent probability. \textbf{Summary Head} exploits the local-GCN module to produce the shot-level features, which will be used to predict the shot selection probability. 
}.
\label{fig:model-overview}
\end{figure*}

\subsection{Unified and Interactive User Intent}
\subsubsection{Problem Setting}
We introduce a novel problem setting with our proposed unified and interactive intent.
The canonical setting for query-focused video summarization is to output a representative and concise subset of video shots based on the inputting video $\boldsymbol{v}$ of $T$ shots and text query $q_t$. 
We re-define the task by generalizing the text query $q_t$ into the generic query $q$.
Then, we propose to predict not only a final video summary, but also a unified and interactive user intent $\zeta$ for the multi-modality queries. $\zeta$ can be learned implicitly like a latent variable. 
We assume that there are a set of basis intents as $Z =\{ \zeta_1, \zeta_2, ..., \zeta_k \}$ and the user intent $\zeta$ is chosen from the basis intents according to a categorical distribution conditioned on the query $q$ as $\zeta \sim p(\zeta | q, v)$. 
Given the user query $q$, the distribution $p(\zeta | q, v)$ is parameterized by the probability vector of basis intents, $\boldsymbol{p(\zeta)} = [p(\zeta_1 | q,\boldsymbol{v}), p(\zeta_2 | q,\boldsymbol{v}),...,p(\zeta_k | q,\boldsymbol{v})]^T$. 


In practice, the query can be either textual, visual, or other formats. In this work, we only implement the models for textual and visual queries. Following the previous works\cite{sharghi_query-focused_2017}, we represent the text query by two text concepts as $q_t = \{ c_1, c_2 \}$,
where $c_1$, $c_2$ are two concepts.
By comparison, we represent the visual query by a set of representative shots in the original video as $q_v = \{ {u}_1, u_2,..., u_P  \}$ where $P$ is a constant number.

Then, for each shot $s$, we denote $\eta_{s} \in \{True, False\}$ as whether $s$ should be selected in the summarization. We assume that $\eta_s$ is sampled from a Bernoulli distribution conditioned on the intent as 
\begin{equation}
    p(\eta_{s})=p(\eta_{s}|\zeta, \boldsymbol{v}).
    \label{eq:p_score}
\end{equation}
Finally, we can condition the shot selection probability $p(\eta_s)$ on the user query as
\begin{equation}
 p(\eta_s | q) = \Sigma_{i=1}^{i \leq k} p(\zeta_i | q,\boldsymbol{v}) * p(\eta_s | \zeta_i,\boldsymbol{v}).  \label{eq:overall}
\end{equation}
Instead of the deterministic intent $\zeta$, we characterize the user's needs by the distribution $p(\zeta | q,\boldsymbol{v})$, which weights different basis intents as Equation \ref{eq:overall} shows. Such a notion follows the perspective of Bayesianism as the latent variable (intent) is a random variable instead of a deterministic value. The user can iteratively adjust the probability vector $\boldsymbol{p(\zeta)}$ to fine-tune its intent.

Since the shot selection probability is often viewed as a summarization score when $\eta=True$, we use the shot score and selection probability interchangeably in this paper.
To implement Equation \ref{eq:p_score}, and \ref{eq:overall}, we design two modules $\boldsymbol {p}(\zeta|q) = g(q, \boldsymbol{v}: \theta_g)$ (intent module) and $p(\eta_{s}|\zeta) = h(\zeta, \boldsymbol{v}: \theta_h)$ (summary module), where $\theta_g$ and $\theta_h$ are the parameters of $g$ and $h$
\begin{equation}
    p(\eta_{s}|q,\boldsymbol{v}) = \Sigma_{i=1}^{i \leq k} g_i(q, \boldsymbol{v}: \theta_g) * h(\zeta_i, \boldsymbol{v}: \theta_h).
\label{eq:implementation}
\end{equation}
Given the ground truth labels, we can optimize the parameters $\theta_g, \theta_h$ of our modules by the BCE Loss as 
\begin{equation}
    \mathscr{L}_{BCE}(\theta_g, \theta_h) = \Sigma_{t=1}^{t \leq T} log (p(y_t | q, \boldsymbol{v})),
\end{equation}
where $y_t$ is the ground truth label for the $t^{th}$ shot.
\subsubsection{Non-Linear Activation}
The Equation \ref{eq:implementation} strictly follows the selection probability's theoretical definition in Equation \ref{eq:overall}. However, it restricts the capacity of the intent module because the resulting probability is simply the linear combination of $h(\zeta_i, \boldsymbol{v}) * g_i(q, \boldsymbol{v}) $. To address the issue, we trade off the strictness for better performance by adding a non-linearity layer on every basis intent score. Specifically, we employ shifted ReLU\cite{agarap2018deep} as the non-linearity activation. 
\begin{equation}
    p(\eta_{s}|q) = \Sigma_{i=1}^{i \leq k}  ReLU(g_i(q, \boldsymbol{v}) *  h(\zeta_i, \boldsymbol{v}) - \delta ),
\end{equation}
where $\delta$ refers to the threshold value for the shifted ReLU. 





\subsection{GSE-GCN: Granularity-Scalable Ego-Graph Convolutional Networks}
As the shared backbone of intent module $g$ and summary module $h$, the \abbrModel \space exploits two newly proposed components, i.e., Granularity-Scalable Pathways (GS-Pathways) and Ego-Graph Convolutional Network (E-GCN),  to better deal with the temporal multi-granularity and sparsity of correlation respectively. 

\subsubsection{Granularity-Scalable Pathways (GS-Pathways)}
\input{tables/slow-fast-conv}

\begin{figure}[t]
\centering
\includegraphics[width=0.4\textwidth]{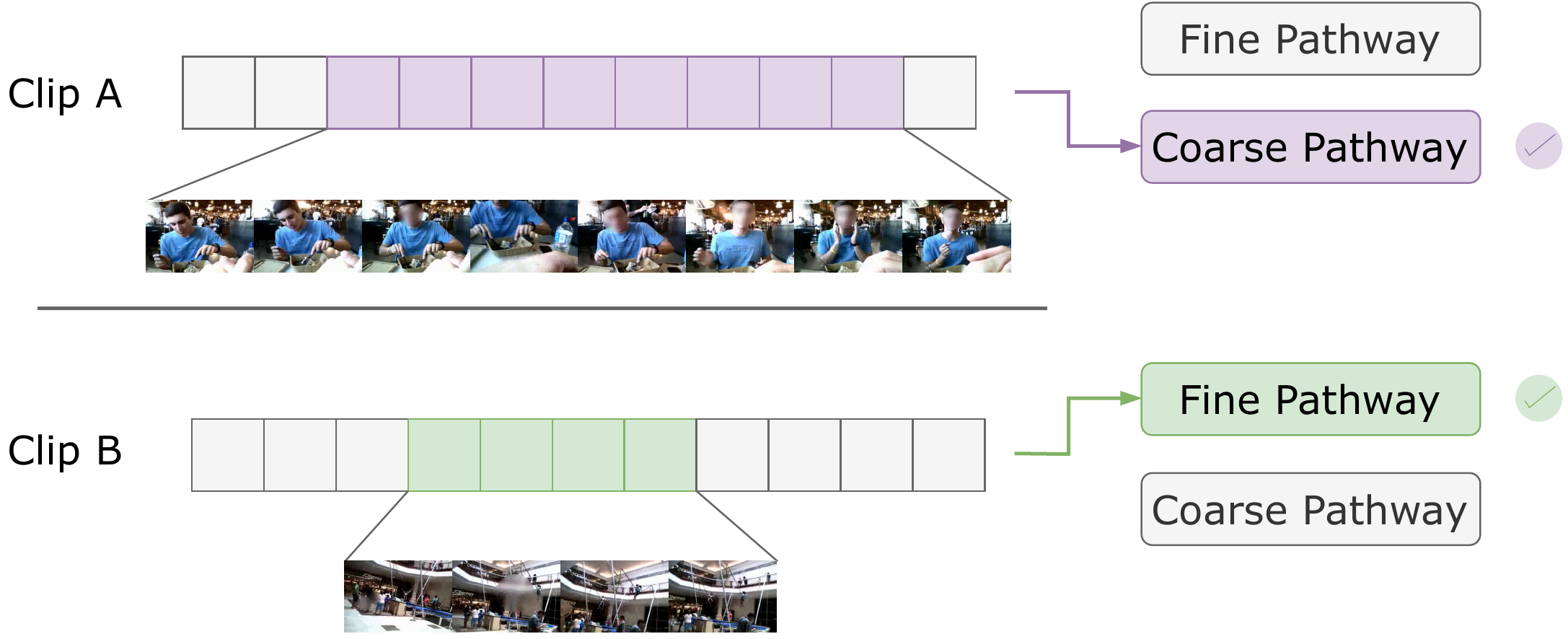} 
\caption{ 
The eating action (Clip A) with a longer length should be processed with the coarser-grained features by the coarse pathway. By comparison, the jumping event (Clip B) with far faster movement should be processed with the finer-grained features.
}
\label{fig:pathway}
\end{figure}
Models with the constant temporal granularity may fall short in aligning the video events/actions of multi-granularity with the user query/intent.
We have shown in Figure \ref{fig:pathway} that the actions of different temporal lengths and movement speeds should be processed with the features of different temporal granularity.
The issue raises the necessity of a granularity-scalable model. To realize it, we propose a flexible structure with two pathways of different granularity. The idea is similar with \cite{feichtenhofer2019slowfast} technically while being motivated by different concerns. For each pathway, we aggregate shot-level features into segment-level features (a segment spans 4 and 16 shots with the fine and coarse pathways, respectively) by a convolutional network. We list the hyper-parameters in Table \ref{tab:slow-fast}. The produced segment-level features are fed into our E-GCN described below to align with the query/intent.

\subsubsection{Ego-Graph Convolutional Networks}
The correlations between the different video segments and query/intent can be relatively sparse given a long video. For example, if the user queries \textit{``walking"}, there can be only a fraction of video content correlated with walking people. Besides, the query-related video content can also have a sparse relationship with other video segments, especially those having a long temporal distance. Thus, correlating all the video segments (e.g., transformer-based models) can be time-inefficient and space-inefficient. We propose to exploit the notion of dynamic edge convolution \cite{wang2019dynamic} and construct a graph $\mathscr{G}=(\mathscr{V},\mathscr{E})$ by connecting the video segments and query/intent dynamically. 
The graph's vertices $\mathscr{V}$ comprises of video segments extracted from the above GS-Pathways and the query/intent. To dynamically model the correlations between the video segments and the user intent, we connect them with the edge set $\mathscr{E}$ consisting of three types of edges, i.e., intent edge $\mathscr{E}_i$, semantic edge $\mathscr{E}_s$, and temporal edge $\mathscr{E}_t$.

\noindent \textbf{Intent Edge} connects the segment vertices with the centric intent vertex, which is why we call the graph as Ego-Graph. We map the intent embedding and segment feature into a mutual latent space dynamically by two MLPs. Then we can derive the intent-segment edge set $\mathscr{E}_z$ as,
\begin{equation}
    \mathscr{E}_z = \{ (w', w_t), w_t \in \mathscr{V}_T  \},
\end{equation}
where $\mathscr{V}_T$ refers to the vertex set of the video segments and $w'$ refers to the mapped query/intent vertex.

\noindent
\textbf{Semantic Edge} connects the video segments with the correlated semantics. Motivated the sparsity of correlation, we follow \cite{xu2020g} and connect the top-k related vertices for each video segment vertex in $\mathscr{V}_T$.
\begin{equation}
    \mathscr{E}_s = \{ (w_t, w_{n_t(k)}) | t = 1,2,..., T; k = 1,2,...,K \},
\end{equation}
where $w_{n_t(k)}$ is the $k^{th}$ nearest neighbor of the vertex $w_t$ in the feature space and $K$ is a constant number.

\noindent
\textbf{Temporal Edge} connects the edges temporally adjacent. Each vertex has a forward edge to the next vertex and a backward edge to the last vertex except the two ends of the segment sequence. We represent the two sets of edges as:
\begin{equation}
    \mathscr{E}^f_t = \{ (w_t, w_{t+1} | t = 1, 2, ..., T-1\},
\end{equation}
\begin{equation}
    \mathscr{E}^b_t = \{ (w_t, w_{t-1} | t = 2,3,..., T\},
\end{equation}
where $\mathscr{E}^f_t$ includes the forward temporal edges, $\mathscr{E}^b_t$ includes the backward temporal edges and $\mathscr{E}_t = \mathscr{E}^b_t \cup \mathscr{E}^b_t$.

\noindent
\textbf{Edge Convolution}
After obtaining the graph, We apply edge convolution as our graph convolution  operation\cite{wang2019dynamic}. 
Following Xu et al.\cite{xu2020g}, We employ the convolution operation to perform the efficient edge convolution on the obtained graph.




\subsubsection{Local Graph for Shot-Level Features}
The output features of edge convolution are at segment-level. 
To reconstruct the shot feature sequence from the segment features, We build the local Ego-Graph for each segment. The graph consists of one segment feature vertex connected with the all spanned shot vertices. We also add the semantic and temporal edges to the graph. 
After applying edge convolution on the constructed graph, we can obtain a shot-level feature sequence.

\subsubsection{Implementation of the Modules} 
Both intent and summary modules are implemented based on \abbrModel \space with different inputs and outputs. The summary module performs element-wise multiplication on the intent embedding and the Local-GCN-processed shot features to get a similarity vector. Then it exploits an MLP with Sigmoid activation to generate the selection probability of shots. By comparison, the intent module exploits an MLP head with Softmax to generate the intent distribution. Since the intent module is designed for the queries of different modalities, there is a slight difference between the visual-query and textual-query. The intent module for textual query strictly follows the GSE-GCN structure, while the intent module for the visual query models the query shots as individual vertices instead of one merged vertex. 

%% file: tables/slow-fast-conv.tex
\begin{table*}[ht]
\centering

\begin{tabular}{c|cccc|cccc}
\multirow{2}{*}{Layer} & \multicolumn{4}{c|}{Coarse Pathway}                                & \multicolumn{4}{c}{Fine Pathway}                                \\
                       & Kernel & Stride & \multicolumn{1}{l}{Channel} & Output Size      & Kernel & Stride & \multicolumn{1}{l}{Channel} & Output Size     \\ \hline
Conv1                  & 5      & 8      & 1024                         & {[}L//8, 1024{]}  & 5      & 1      & 256                         & {[}L//2, 256{]} \\ \hline
MaxPool1               & 2      & 1      & 1024                         & {[}L//8, 1024{]}  & 2      & 2      & 256                         & {[}L//2, 256{]} \\ \hline
Conv2                  & 5      & 1      & 1024                         & {[}L//8, 1024{]}  & 5      & 1      & 256                         & {[}L//2, 256{]} \\ \hline
MaxPool2               & 3      & 2      & 1024                         & {[}L//16, 1024{]} & 2      & 2      & 256                         & {[}L//4, 256{]}
\end{tabular}
\caption{The hyperparameter setting of the granularity-scalable pathways. $L$ refers to the length of the original video.}
\label{tab:slow-fast}
\end{table*}

%% file: sections/experiments.tex
\section{Experiments}

\newcommand{\ours}{\textbf{53.58} & \textbf{53.27} & \multicolumn{1}{l}{\textbf{50.90}}}

\begin{figure*}[t]
\centering
\includegraphics[trim=0 0 0 95,clip, width=0.88\textwidth]{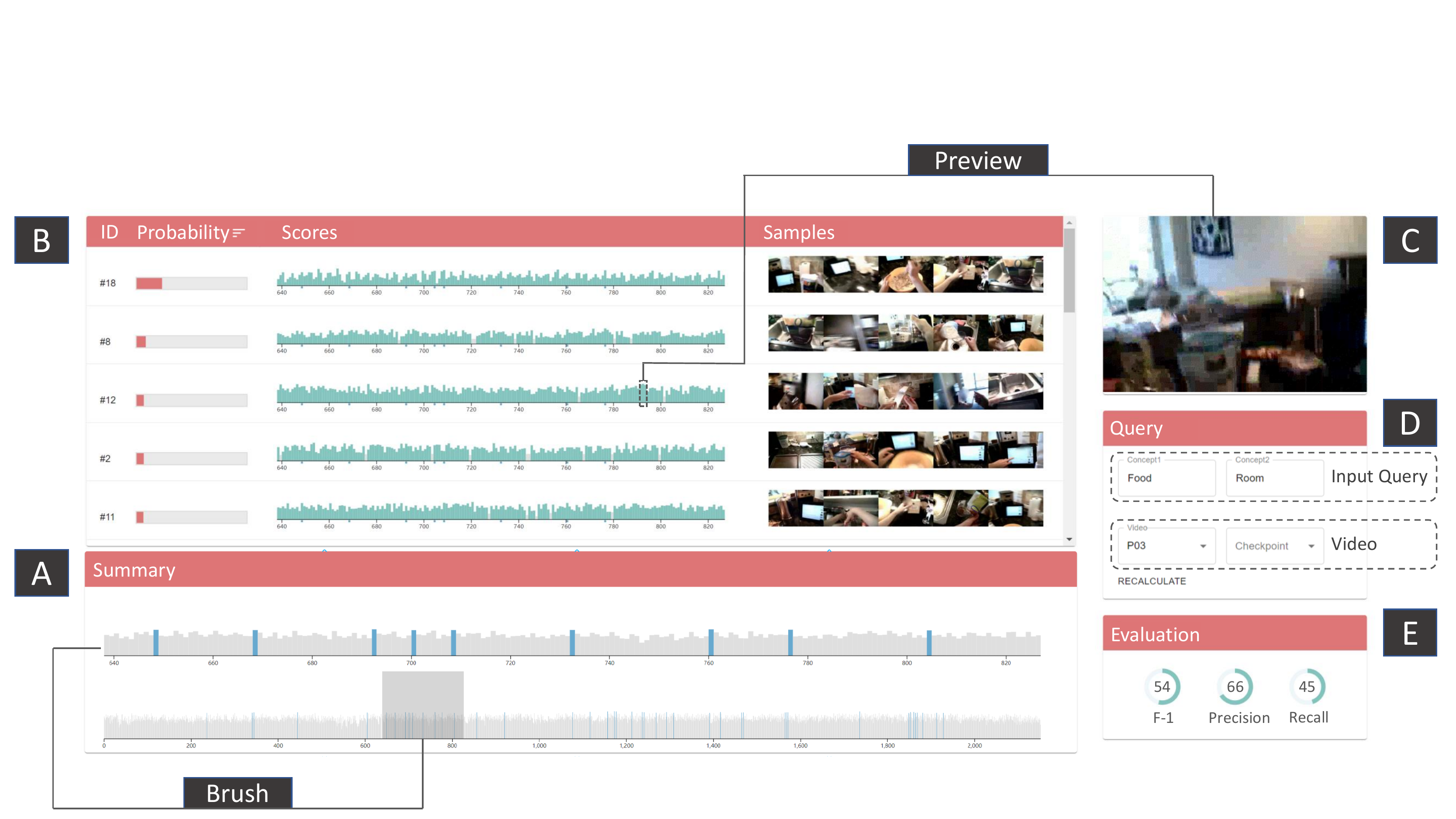} 
\caption{Prototype Overview. \textbf{A}: \textbf{Summary View} presents two temporal bar charts, which shows the overall scores and the summarized shots. The bottom bar chart shows the overview of all the shots, while the top bar chart zooms into the detail decided by the brush in the bottom chart. \textbf{B}: \textbf{Intent View} list all the basis intents with their probability, shot scores, and representative samples. The samples are selected with the highest score. \textbf{C}: \textbf{Preview View} plays a GIF of the user-hovering shot. In this case, the user hovers on the highlighted shot in intent \#12, which includes a room scenario. \textbf{D}: \textbf{Query View} allows the user to change the query and makes the model rerun. \textbf{E} \textbf{Evaluation View} shows the quantitative result of the summary.}
\label{fig:prototype-overview}
\end{figure*}
\input{tables/text-comparisom}
\input{tables/visual-comparison}

\subsection{Implementation Detail}
We exploit Pytorch \cite{paszke2019pytorch} to implement our model on an NVIDIA RTX 8000. 
We have 20 basis intents, each represented by a 128-D embedding vector.
For the summary module, we employ a 3-layer GCN and a 3-layer MLP after the GCN layers. We set the number of GCN layers and MLP layers as 2 and 3 for the intent module. 
Models are trained by an Adam optimizer with a base learning rate of 1e-4. We employ a warm-up strategy \cite{goyal2017accurate} to linearly increase the learning rate from 0 to the base learning rate in 10 epochs. After that, we reduce the learning rate to one-tenth of the previous value every twenty epochs. 


\subsection{Experiment Setting}

\subsubsection{Text Query Dataset}
We conduct our text-query experiments on the query-driven video summarization dataset \cite{sharghi_query-focused_2017}. The dataset includes the four videos in UT Egocentric(UTE) dataset\cite{lee2012discovering}.  Each of the videos (3-5 hours) is captured in daily life scenarios. Each query in the dataset is represented by two concepts among the total 48 concepts. 
\subsubsection{Visual Query Dataset and Dataset Baselines}
We build our visual-query dataset based on the text-query dataset. For each annotated summarization, we employ the eigenvector centrality as the criteria to pick the most representative shots as the query shots. Detailed examples and descriptions can be found in the supplementary materials.
As the visual query dataset is newly introduced and no previous work exists, we compare our approach with two baselines, i.e., linear prediction and attentional query model, which can be founded in the supplementary.




\subsubsection{Evaluation Protocol}
To compare with the previous approaches fairly, we employ the semantic evaluation protocol proposed by Sharghi et al.\cite{sharghi_query-focused_2017}. This protocol is based on the semantic similarity between the machine-generated and the ground-truth video shots. The similarity is generated through finding the maximum weight matching of the bipartite graph computed based on the semantic interception-over-union (IOU). The summed weights of the matched shot pairs are used to compute precision, recall, and F-1 measure. 
Note that for the visual query dataset, we mask out the query shots in the evaluation stage. 
To evaluate the interactive intent, which involves the human subjectivity, we develop a prototype and propose a case study in qualitative analysis.

\subsection{Comparative Analysis}
The comparison among our method and SOTA methods for the Textual Query Task can be found in Table \ref{tab:text-comparison}. We observe that our method achieves the highest F-1 value of 50.90\%. 
The result proves that our method can better identify the correlation between the query and summary. We also present the experimental result on the visual query task in Table \ref{tab:visual-comparison}. We find our method significantly outperforms the baselines by 7\%, although the general performance is inferior to the performance on the text query task.

\subsection{Ablation Analysis}
We evaluate the effects of the proposed methods and find the optimum model setting by an ablation study. 

\subsubsection{Ego-Graph Convolutional Networks}
\input{tables/ablation-gcn}
Our proposed Ego-GCN approach aligns the video segment features with the query/intent. To validate the effectiveness, we replace Ego-GCN by the transformer-based approach\cite{su2019vl, sun2019videobert, zhou2020unified} in the intent and summary modules iteratively.
The experiment results can be found in Table \ref{tab:ablation-gcn}. Our Ego-GCN can advance the model performance when added to either intent or summary module.

\subsubsection{Local GCN}
\input{tables/ablation-upsample}
We employ Local GCN to recover the shot-level features from the segment-level features. As shown in Figure \ref{tab:ablation-upsampling}, Local GCN's performance is superior to bi-cubic upsampling and transpose convolutional layer, which is used in \cite{xiao_convolutional_2020}. 

\subsubsection{GS-Pathway}
\input{tables/ablation-pathway}
To validate the effects of our GS-Pathway, we compare our model with three variants with only one fixed pathway. 
We present the experiment results in Table \ref{tab:ablation-pathway}. Our result shows our model surpasses the three variants, verifying the necessity of attending the segment features of multi-granularity.

\subsubsection{When to fusion the pathways?}
\input{tables/fusion}
The features of different pathways can fuse at different stages. To find the optimum of the model, we compare the variants with different fusion strategies, i.e, early, middle, late fusions. 
The early fusion strategy fusions the features before the dot product shown in Fig. \ref{fig:model-overview}. The middle fusion and late fusion happen before and after the MLP, respectively.
As can be found in Table \ref{tab:fusion}, fusion at the early stage is the best choice.

\subsubsection{Do we need video as input of the intent modules?}
\input{tables/query-model}
Our intent module use both the query and video as input to predict the user intent. 
However, the user intent can also solely rely on the user query, as some users might not have time to browse the original videos. Thus, it is necessary to learn if we can remove the video inputs from the intent module and let it infer only based on the user query.
To do so, we compare the full model with two variants using a simpler intent module and a video-agnostic intent module. We present the experiment result in Table \ref{tab:intent-video}. Though there is a marginal performance decrease, the model with a video-agnostic intent module still outperforms most of the previous state-of-the-art approaches. The result shows it is reasonable to remove the video input for the intent module to promote the model's generalizability.
\subsubsection{Can we transfer the summary module between different datasets?}
\input{tables/transferring}

To validate the generality of the summary module, we experiment on the visual-query task in the transferring setting. We first train the summary module on the text-query dataset. Then, we reuse this pre-trained summary module and only train the intent module for the visual-query task. The results can be found in Table \ref{tab:transferring}. The experiment result shows that the transferred model surpasses the model trained in the canonical setting, showing that the summary module is interchangeable for the queries of multi-modality. 

\subsection{Prototype and Qualitative Analysis}
We demonstrate the interactivity of our framework with a prototype as shown in Fig. \ref{fig:prototype-overview}. The prototype can also work as qualitative analysis, to prove that our approach can generate the query-related summary with better interpretability. In the figure, we show an example case. The snapshot is taken when the user queries ``Food" and ``Room" for video-3. Note here we set that the user input is always two queries in our design of the prototype, although the proposed framework can deal with other numbers of queries.  The user first brushes on the result view and focuses on the clip where more shots are captured in the summary.
Based on the Intent View (B), we can identify intent \#18, \#8, \#12, \#2, and \#11 in descending order. From the samples of each intent, we find the \#intent \#18 and \#8 are closely related to the food cooking scenarios when \#11 contains some food storage scenarios. The \#12 and \#2 are more likely to focus on the room scenarios. We also observe that there are some computer frames in \#12 and \#2. Previewing the related shot, we notice that the computer is the foreground object of the room, as Fig. \ref{fig:prototype-overview}. C shows. Thus, the snapshot shows that our model successfully captures the food and room scenarios. We can find that there are two types of food scenarios identified, i.e., food cooking and food storage. This finding also shows that our approach can provide finer-grained user intent representation. 

%% file: tables/text-comparisom.tex
\begin{table*}[!htbp]
\centering
\resizebox{\textwidth}{!}{%
\begin{tabular}{|l|lll|lll|lll|lll|lll|}
\hline
\multicolumn{1}{|c|}{} & \multicolumn{3}{c|}{Video-1} & \multicolumn{3}{c|}{Video-2} & \multicolumn{3}{c|}{Video-3} & \multicolumn{3}{c|}{Video-4} & \multicolumn{3}{c|}{Avg.} \\ \cline{2-16} 
\multicolumn{1}{|c|}{\multirow{-2}{*}{Method}} & \multicolumn{1}{c}{Pre.} & \multicolumn{1}{c}{Rec.} & \multicolumn{1}{c|}{F-1} & \multicolumn{1}{c}{Pre.} & \multicolumn{1}{c}{Rec.} & \multicolumn{1}{c|}{F-1} & \multicolumn{1}{c}{Pre.} & \multicolumn{1}{c}{Rec.} & \multicolumn{1}{c|}{F-1} & \multicolumn{1}{c}{Pre.} & \multicolumn{1}{c}{Rec.} & \multicolumn{1}{c|}{F-1} & \multicolumn{1}{c}{Pre.} & \multicolumn{1}{c}{Rec.} & \multicolumn{1}{c|}{F-1} \\ \hline
QC-DPP \cite{sharghi_query-focused_2017} & 49.86 & 53.38 & 48.68 & 33.71 & 62.09 & 41.66 & 55.16 & 29.24 & 36.51 & 21.39 & 63.12 & 29.96 & 40.03 & 60.25 & 44.19 \\ \hline
CHAN \cite{xiao_convolutional_2020} & 54.73 & 46.57 & 49.14 & 45.92 & 50.26 & 46.53 & 59.75 & 64.53 & 58.65 & 25.23 & 51.16 & 33.42 & 46.40 & 53.13 & 46.94 \\ \hline
HVN \cite{jiang_hierarchical_2019} & 52.55 & 52.91 & 51.45 & 38.66 & 62.70 & 47.49 & 60.28 & 62.58 & 61.08 & 26.27 & 54.21 & 35.47 & 44.57 & \textbf{58.10} & 48.87 \\ \hline
QSAN \cite{xiao_query-biased_2020}& 48.41 & 52.34 & 48.52 & 46.51 & 51.36 & 46.64 & 56.78 & 61.14 & 56.93 & 30.54 & 46.90 & 34.25 & 45.56 & {52.94} & 46.59 \\ \hline
Nalla et al. \cite{nalla_watch_nodate} & 54.58 & 52.51 & 50.96 & 48.12 & 52.15 & 48.28 & 58.48 & 61.66 & 58.41 & 37.40 & 43.90 & 39.18 & \underline{49.64} & 52.55 & \underline{49.20} \\ \hline
Ours & \multicolumn{1}{r}{{62.19}} & \multicolumn{1}{r}{{45.23}} & \multicolumn{1}{r|}{{51.27}} & \multicolumn{1}{r}{{50.43}} & \multicolumn{1}{r}{{57.81}} & \multicolumn{1}{r|}{{53.48}} & \multicolumn{1}{r}{{73.45}} & \multicolumn{1}{r}{{53.56}} & \multicolumn{1}{r|}{{61.58}} & \multicolumn{1}{r}{{28.24}} & \multicolumn{1}{r}{{56.47}} & \multicolumn{1}{r|}{{37.25}} & \multicolumn{1}{r}{{\textbf{53.58}}} & \multicolumn{1}{r}{{\underline{53.27}}} & \multicolumn{1}{r|}{{\textbf{50.90}}} \\ \hline
\end{tabular}%
}
\caption{Textual Query Dataset: Comparison with the previous state-of-the-art approaches.}
\label{tab:text-comparison}
\end{table*}

%% file: tables/visual-comparison.tex
\begin{table*}[!htbp]
\centering
\resizebox{\textwidth}{!}{%
\begin{tabular}{|l|lll|lll|lll|lll|lll|}
\hline
\multicolumn{1}{|c|}{} & \multicolumn{3}{c|}{Video-1} & \multicolumn{3}{c|}{Video-2} & \multicolumn{3}{c|}{Video-3} & \multicolumn{3}{c|}{Video-4} & \multicolumn{3}{c|}{Avg.} \\ \cline{2-16} 
\multicolumn{1}{|c|}{\multirow{-2}{*}{Method}} & \multicolumn{1}{c}{Pre.} & \multicolumn{1}{c}{Rec.} & \multicolumn{1}{c|}{F-1} & \multicolumn{1}{c}{Pre.} & \multicolumn{1}{c}{Rec.} & \multicolumn{1}{c|}{F-1} & \multicolumn{1}{c}{Pre.} & \multicolumn{1}{c}{Rec.} & \multicolumn{1}{c|}{F-1} & \multicolumn{1}{c}{Pre.} & \multicolumn{1}{c}{Rec.} & \multicolumn{1}{c|}{F-1} & \multicolumn{1}{c}{Pre.} & \multicolumn{1}{c}{Rec.} & \multicolumn{1}{c|}{F-1} \\ \hline
Linear Baseline & 
59.24 & 45.33 & 49.75 & 21.49 & 26.71 & 23.62 & 56.09 & 44.42 & 49.22 & 14.44 & 33.1 & 19.77 & 37.82 & 37.39 & 35.59 \\ \hline
Attention Baseline &45.01 & 33.96 & 37.71 & 38.86 & 48.01 & 41.09 & 57.7 & 48.75 & 50.66 & 18.00 & 41.5 & 24.75 & 39.89 & 43.06 & 38.55 \\ \hline
Ours & 58.17 & 44.91 & 49.43 & 42.52 & 52.69 & 46.64 & 65.45 & 51.92 & 57.49 & 21.15 & 49.23 & 29.19 & \textbf{46.82} & \textbf{49.69} & \textbf{45.69} \\ \hline

\end{tabular}%
}
\caption{Visual Query Dataset: Comparison with the baselines.}
\label{tab:visual-comparison}
\end{table*}

%% file: tables/ablation-gcn.tex
\begin{table}[!htbp]
\centering
\begin{tabular}{l|l|ccc}
\hline
I.M. & S.M.  & Pre. & Rec. & F-1 \\ \hline
Transformer & Transformer & 44.82 &	44.52 &	42.68 \\ \hline
Transformer & Ego-GCN &   49.00 & 47.89 & 46.15  \\ \hline
Ego-GCN & Transformer &  47.09 & 47.26	& 44.75 \\ \hline
Ego-GCN & Ego-GCN & \ours \\ \hline
\end{tabular}%
\caption{Ablation study the proposed Ego-GCN. I.M. refers to the intent module when S.M. refers to the summary module.}
\label{tab:ablation-gcn}
\end{table}

%% file: tables/ablation-upsample.tex
\begin{table}[!htbp]
\centering

{

\begin{tabular}{l|ccc}
\hline
Pathway & Pre. & Rec. & F-1 \\ \hline
Upsampling & 38.04 & 37.48 & 35.88\\ \hline
Transpose Conv & 47.53 & 47.41 & 45.18 \\ \hline
Local GCN & \textbf{53.58} & \textbf{53.27} & \multicolumn{1}{l}{\textbf{50.90}}  \\ \hline

\end{tabular}%

}
\caption{Ablation study for Local GCN.}
\label{tab:ablation-upsampling}
\end{table}

%% file: tables/ablation-pathway.tex
\begin{table}[!htbp]
\centering
{%
\begin{tabular}{l|ccc}
\hline
Pathway & Pre. & Rec. & F-1 \\ \hline
Shot-Level Feature & 47.45 & 45.38 & 44.40 \\ \hline
Coarse-Granularity & {47.40} & {47.66} & {45.15} \\ \hline
Fine-Granularity & {50.18} & {50.23} & {47.81} \\ \hline
Full-Model & \ours  \\ \hline
\end{tabular}%

}
\caption{Ablation study for the GS-Pathway. The pathway \textit{Shot-Level Feature} refers to the model that directly applies  our Ego-GCN on the shot-level video features.}
\label{tab:ablation-pathway}
\end{table}

%% file: tables/fusion.tex
\begin{table}[!htbp]
\centering
{%
\begin{tabular}{l|ccc}
\hline
Stage & Pre. & Rec. & F-1 \\ \hline
Early &  \ours \\ \hline
Middle & 49.53  & 48.66  & 46.66 \\ \hline
Late & 47.69  & 47.98 & 45.47 \\ \hline
\end{tabular}%
}
\caption{Experiment result on the feature fusion stage.}
\label{tab:fusion}
\end{table}

%% file: tables/query-model.tex
\begin{table}[!htbp]
\centering
\begin{tabular}{l|ccc}
\hline
Intent Module & Pre. & Rec. & F-1 \\ \hline
Video Agnostic & 50.06  & 48.78  & 47.15 \\ \hline
Query Attention &  49.26 & 47.85 & 46.27 \\ \hline
Full Model & \ours \\ \hline
\end{tabular}%
\caption{Experiments on the video inputs in the intent module.}
\label{tab:intent-video}
\end{table}

%% file: tables/transferring.tex
\begin{table}[!htbp]
\centering
{%
\begin{tabular}{l|ccc}
\hline
Training & Pre. & Rec. & F-1 \\ \hline
Canonical & 46.82 & 49.69 & 45.69  \\ \hline
Transferring & \textbf{47.15}  & \textbf{51.08}  & \textbf{46.40} \\ \hline
\end{tabular}%


}
\caption{Experiment on visual query task with transfer learning.}
\label{tab:transferring}
\end{table}

%% file: sections/conclusion.tex
\section{Discussion}
\subsection{Limitations}
\textbf{Adaptive Granularity}: 
Though our ablation study prove the necessity of the granularity-scalable method in video summarization, our proposed approach is not memory-efficient with two pathways of fixed granularity. We will design an adaptive way to control the temporal granularity of the video segments with lower space complexity.

\textbf{More Modalities}: Our IntentVizor framework can support the queries of different modalities. However, we only evaluate our approach on textual and visual queries. Our future work will extend our framework to support other queries, e.g., audio, sketch, etc.


\textbf{Evaluation of the Interface}:
Despite that our proposed interface can assist users in the video summarization by a case study, the effectiveness of this interface should be also verified by an in-the-wild user study in the future. 

\subsection{Applications}
As our proposed IntentVizor improves the interpretability and interactivity of the video summarization, it also has potential practical value. As the user can control the output of summary adaptively based on their detailed needs, such summarization can widely be used in customer-obsessed video browsing, data transportation/recording, surveillance analysis, and sports game highlights etc. 
\subsection{Conclusion}
In this work, we propose IntentVizor, an interactive video summarization framework guided by the generic query. First, our framework introduces a novel concept ``intent", which originally come from Information Retrieval (IR) community, to represent the multi-modality queries. Second, we develop a prototype to make the proposed framework interactive with the user. The user can control the intent to generate summaries satisfying their needs. Third, for the model part, two novel intent/summary modules are designed to better understand the generic queries and generate summaries accordingly/adaptively. Both quantitative and qualitative experiment results verify the superiority of our proposed approach. Four ablation studies also verify more potential extensibility of the proposed framework. For future work, we will solve the limitation above, and introduce more query modalities to better satisfy users' video summarization needs.

%% file: sections/acknowledgement.tex
\section{Acknowledgement}
Guande Wu was partially supported by an NYU School of Engineering Fellowship. This research is also partially funded  by  C2SMART,  a  Tier  1  University  Center  awarded  by  U.S.  Department  of  Transportation under contract 69A3351747124, and NSF awards CNS-1229185, CCF-1533564, CNS-1544753, CNS-1730396, and CNS-1828576. Lin, Silva, and Wu are partially funded by DARPA Perceptually-enabled Task Guidance (PTG). Any opinions, findings, and conclusions or recommendations expressed in this material are those of the authors and do not necessarily reflect the views of NSF, USDOT, or DARPA.

%% file: cvpr.bbl
\begin{thebibliography}{10}\itemsep=-1pt

\bibitem{agarap2018deep}
Abien~Fred Agarap.
\newblock Deep learning using rectified linear units (relu).
\newblock {\em CoRR}, abs/1803.08375, 2018.

\bibitem{apostolidis2020ac}
Evlampios Apostolidis, Eleni Adamantidou, Alexandros~I Metsai, Vasileios
  Mezaris, and Ioannis Patras.
\newblock Ac-sum-gan: Connecting actor-critic and generative adversarial
  networks for unsupervised video summarization.
\newblock {\em IEEE Transactions on Circuits and Systems for Video Technology},
  2020.

\bibitem{broder2002taxonomy}
Andrei Broder.
\newblock A taxonomy of web search.
\newblock In {\em ACM SIGIR Forum}, volume~36, pages 3--10, 2002.

\bibitem{broder2007robust}
Andrei~Z Broder, Marcus Fontoura, Evgeniy Gabrilovich, Amruta Joshi, Vanja
  Josifovski, and Tong Zhang.
\newblock Robust classification of rare queries using web knowledge.
\newblock In {\em Proceedings of the 30th International ACM SIGIR Conference on
  Research and Development in Information Retrieval}, pages 231--238, 2007.

\bibitem{chu2015video}
Wen-Sheng Chu, Yale Song, and Alejandro Jaimes.
\newblock Video co-summarization: Video summarization by visual co-occurrence.
\newblock In {\em Proceedings of the IEEE Conference on Computer Vision and
  Pattern Recognition}, pages 3584--3592, 2015.

\bibitem{del2017active}
Ana~Garcia del Molino, Xavier Boix, Joo-Hwee Lim, and Ah-Hwee Tan.
\newblock Active video summarization: Customized summaries via on-line
  interaction with the user.
\newblock In {\em Proceedings of the AAAI Conference on Artificial
  Intelligence}, volume~31, 2017.

\bibitem{fajtl2018summarizing}
Jiri Fajtl, Hajar~Sadeghi Sokeh, Vasileios Argyriou, Dorothy Monekosso, and
  Paolo Remagnino.
\newblock Summarizing videos with attention.
\newblock In {\em Proceedings of the Asian Conference on Computer Vision},
  pages 39--54, 2018.

\bibitem{fan2019understanding}
Lifeng Fan, Wenguan Wang, Siyuan Huang, Xinyu Tang, and Song-Chun Zhu.
\newblock Understanding human gaze communication by spatio-temporal graph
  reasoning.
\newblock In {\em Proceedings of the IEEE/CVF International Conference on
  Computer Vision}, pages 5724--5733, 2019.

\bibitem{feichtenhofer2019slowfast}
Christoph Feichtenhofer, Haoqi Fan, Jitendra Malik, and Kaiming He.
\newblock Slowfast networks for video recognition.
\newblock In {\em Proceedings of the IEEE/CVF International Conference on
  Computer Vision}, pages 6202--6211, 2019.

\bibitem{ghosh2020stacked}
Pallabi Ghosh, Yi Yao, Larry Davis, and Ajay Divakaran.
\newblock Stacked spatio-temporal graph convolutional networks for action
  segmentation.
\newblock In {\em Proceedings of the IEEE/CVF Winter Conference on Applications
  of Computer Vision}, pages 576--585, 2020.

\bibitem{goyal2017accurate}
Priya Goyal, Piotr Doll{\'a}r, Ross Girshick, Pieter Noordhuis, Lukasz
  Wesolowski, Aapo Kyrola, Andrew Tulloch, Yangqing Jia, and Kaiming He.
\newblock Accurate, large minibatch sgd: Training imagenet in 1 hour.
\newblock {\em CoRR}, abs/1706.02677, 2017.

\bibitem{guo2016deep}
Jiafeng Guo, Yixing Fan, Qingyao Ai, and W~Bruce Croft.
\newblock A deep relevance matching model for ad-hoc retrieval.
\newblock In {\em Proceedings of the ACM International Conference on
  Information and Knowledge Management}, pages 55--64, 2016.

\bibitem{jiang_hierarchical_2019}
Pin Jiang and Yahong Han.
\newblock Hierarchical variational network for user-diversified \&
  query-focused video summarization.
\newblock In {\em Proceedings of the 2019 International Conference on
  Multimedia Retrieval}, pages 202--206, 2019.

\bibitem{jin2017elasticplay}
Haojian Jin, Yale Song, and Koji Yatani.
\newblock Elasticplay: Interactive video summarization with dynamic time
  budgets.
\newblock In {\em Proceedings of the 25th ACM International Conference on
  Multimedia}, pages 1164--1172, 2017.

\bibitem{jung2019discriminative}
Yunjae Jung, Donghyeon Cho, Dahun Kim, Sanghyun Woo, and In~So Kweon.
\newblock Discriminative feature learning for unsupervised video summarization.
\newblock In {\em Proceedings of the AAAI Conference on Artificial
  Intelligence}, volume~33, pages 8537--8544, 2019.

\bibitem{kanafani2021unsupervised}
Hussain Kanafani, Junaid~Ahmed Ghauri, Sherzod Hakimov, and Ralph Ewerth.
\newblock Unsupervised video summarization via multi-source features.
\newblock In {\em Proceedings of the 2021 International Conference on
  Multimedia Retrieval}, page 466–470, 2021.

\bibitem{kofler2016user}
Christoph Kofler, Martha Larson, and Alan Hanjalic.
\newblock User intent in multimedia search: a survey of the state of the art
  and future challenges.
\newblock {\em ACM Computing Surveys (CSUR)}, 49(2):1--37, 2016.

\bibitem{lee2012discovering}
Yong~Jae Lee, Joydeep Ghosh, and Kristen Grauman.
\newblock Discovering important people and objects for egocentric video
  summarization.
\newblock In {\em Proceedings of the IEEE Conference on Computer Vision and
  Pattern Recognition}, pages 1346--1353, 2012.

\bibitem{liu2021spatial}
Jiawei Liu, Zheng-Jun Zha, Wei Wu, Kecheng Zheng, and Qibin Sun.
\newblock Spatial-temporal correlation and topology learning for person
  re-identification in videos.
\newblock In {\em Proceedings of the IEEE/CVF Conference on Computer Vision and
  Pattern Recognition}, pages 4370--4379, 2021.

\bibitem{mahasseni2017unsupervised}
Behrooz Mahasseni, Michael Lam, and Sinisa Todorovic.
\newblock Unsupervised video summarization with adversarial lstm networks.
\newblock In {\em Proceedings of the IEEE Conference on Computer Vision and
  Pattern Recognition}, pages 202--211, 2017.

\bibitem{mavroudi2019neural}
Effrosyni Mavroudi, Benjam{\'\i}n B{\'e}jar~Haro, and Ren{\'e} Vidal.
\newblock Representation learning on visual-symbolic graphs for video
  understanding.
\newblock In {\em Proceedings of the European Conference on Computer Vision},
  pages 71--90, 2020.

\bibitem{messaoud_deepqamvs_2021}
Safa Messaoud, Ismini Lourentzou, Assma Boughoula, Mona Zehni, Zhizhen Zhao,
  Chengxiang Zhai, and Alexander~G. Schwing.
\newblock {DeepQAMVS}: Query-aware hierarchical pointer networks for
  multi-video summarization.
\newblock In {\em Proceedings of the 44th International ACM SIGIR Conference on
  Research and Development in Information Retrieval}, page 1389–1399, 2021.

\bibitem{nalla_watch_nodate}
Saiteja Nalla, Mohit Agrawal, Vishal Kaushal, Ganesh Ramakrishnan, and Rishabh
  Iyer.
\newblock Watch hours in minutes: Summarizing video with user intent.
\newblock In {\em Proceedings of the European Conference on Computer Vision},
  pages 714--730, 2020.

\bibitem{pan2020spatio}
Boxiao Pan, Haoye Cai, De-An Huang, Kuan-Hui Lee, Adrien Gaidon, Ehsan Adeli,
  and Juan~Carlos Niebles.
\newblock Spatio-temporal graph for video captioning with knowledge
  distillation.
\newblock In {\em Proceedings of the IEEE/CVF Conference on Computer Vision and
  Pattern Recognition}, pages 10870--10879, 2020.

\bibitem{panda2017diversity}
Rameswar Panda, Niluthpol~Chowdhury Mithun, and Amit~K Roy-Chowdhury.
\newblock Diversity-aware multi-video summarization.
\newblock {\em IEEE Transactions on Image Processing}, 26(10):4712--4724, 2017.

\bibitem{papalampidi2020movie}
Pinelopi Papalampidi, Frank Keller, and Mirella Lapata.
\newblock Movie summarization via sparse graph construction.
\newblock In {\em Proceedings of the AAAI Conference on Artificial
  Intelligence}, pages 13631--13639, 2020.

\bibitem{park2020sumgraph}
Jungin Park, Jiyoung Lee, Ig-Jae Kim, and Kwanghoon Sohn.
\newblock Sumgraph: Video summarization via recursive graph modeling.
\newblock In {\em Proceedings of the European Conference on Computer Vision},
  pages 647--663, 2020.

\bibitem{paszke2019pytorch}
Adam Paszke, Sam Gross, Francisco Massa, Adam Lerer, James Bradbury, Gregory
  Chanan, Trevor Killeen, Zeming Lin, Natalia Gimelshein, Luca Antiga, et~al.
\newblock Pytorch: An imperative style, high-performance deep learning library.
\newblock {\em Advances in Neural Information Processing Systems},
  32:8026--8037, 2019.

\bibitem{sharghi_query-focused_2017}
Aidean Sharghi, Jacob~S. Laurel, and Boqing Gong.
\newblock Query-focused video summarization: dataset, evaluation, and a memory
  network based approach.
\newblock In {\em Proceedings of the IEEE {Conference} on {Computer} {Vision}
  and {Pattern} {Recognition}}, pages 2127--2136, 2017.

\bibitem{su2019vl}
Weijie Su, Xizhou Zhu, Yue Cao, Bin Li, Lewei Lu, Furu Wei, and Jifeng Dai.
\newblock Vl-bert: Pre-training of generic visual-linguistic representations.
\newblock In {\em International Conference on Learning Representations}, 2020.

\bibitem{sun2019videobert}
Chen Sun, Austin Myers, Carl Vondrick, Kevin Murphy, and Cordelia Schmid.
\newblock Videobert: A joint model for video and language representation
  learning.
\newblock In {\em Proceedings of the IEEE/CVF International Conference on
  Computer Vision}, pages 7464--7473, 2019.

\bibitem{wang2021weakly}
Mingui Wang, Di Cui, Lifang Wu, Meng Jian, Yukun Chen, Dong Wang, and Xu Liu.
\newblock Weakly-supervised video object localization with attentive
  spatio-temporal correlation.
\newblock {\em Pattern Recognition Letters}, 145:232--239, 2021.

\bibitem{wang2018videos}
Xiaolong Wang and Abhinav Gupta.
\newblock Videos as space-time region graphs.
\newblock In {\em Proceedings of the European Conference on Computer Vision},
  pages 399--417, 2018.

\bibitem{wang2019dynamic}
Yue Wang, Yongbin Sun, Ziwei Liu, Sanjay~E Sarma, Michael~M Bronstein, and
  Justin~M Solomon.
\newblock Dynamic graph cnn for learning on point clouds.
\newblock {\em ACM Transactions on Graphics}, 38(5):1--12, 2019.

\bibitem{xiao_query-biased_2020}
Shuwen Xiao, Zhou Zhao, Zijian Zhang, Ziyu Guan, and Deng Cai.
\newblock Query-biased self-attentive network for query-focused video
  summarization.
\newblock {\em IEEE Transactions on Image Processing}, 29:5889--5899, 2020.

\bibitem{xiao_convolutional_2020}
Shuwen Xiao, Zhou Zhao, Zijian Zhang, Xiaohui Yan, and Min Yang.
\newblock Convolutional hierarchical attention network for query-focused video
  summarization.
\newblock In {\em Proceedings of the AAAI Conference on Artificial
  Intelligence}, volume~34, pages 12426--12433, 2020.

\bibitem{xu2020g}
Mengmeng Xu, Chen Zhao, David~S Rojas, Ali Thabet, and Bernard Ghanem.
\newblock G-tad: Sub-graph localization for temporal action detection.
\newblock In {\em Proceedings of the IEEE/CVF Conference on Computer Vision and
  Pattern Recognition}, pages 10156--10165, 2020.

\bibitem{yaliniz2019unsupervised}
G{\"o}khan Yalınız and Nazli Ikizler-Cinbis.
\newblock Unsupervised video summarization with independently recurrent neural
  networks.
\newblock In {\em 27th Signal Processing and Communications Applications
  Conference (SIU)}, pages 1--4, 2019.

\bibitem{yan2018spatial}
Sijie Yan, Yuanjun Xiong, and Dahua Lin.
\newblock Spatial temporal graph convolutional networks for skeleton-based
  action recognition.
\newblock In {\em Proceedings of the AAAI Conference on Artificial
  Intelligence}, pages 1113--1122, 2018.

\bibitem{yin2010building}
Xiaoxin Yin and Sarthak Shah.
\newblock Building taxonomy of web search intents for name entity queries.
\newblock In {\em Proceedings of the 19th International Conference on World
  Wide Web}, pages 1001--1010, 2010.

\bibitem{zamani2016estimating}
Hamed Zamani and W~Bruce Croft.
\newblock Estimating embedding vectors for queries.
\newblock In {\em Proceedings of the 2016 ACM International Conference on the
  Theory of Information Retrieval}, pages 123--132, 2016.

\bibitem{zeng2019graph}
Runhao Zeng, Wenbing Huang, Mingkui Tan, Yu Rong, Peilin Zhao, Junzhou Huang,
  and Chuang Gan.
\newblock Graph convolutional networks for temporal action localization.
\newblock In {\em Proceedings of the IEEE/CVF International Conference on
  Computer Vision}, pages 7094--7103, 2019.

\bibitem{zhang2019generic}
Hongfei Zhang, Xia Song, Chenyan Xiong, Corby Rosset, Paul~N Bennett, Nick
  Craswell, and Saurabh Tiwary.
\newblock Generic intent representation in web search.
\newblock In {\em Proceedings of the 42nd International ACM SIGIR Conference on
  Research and Development in Information Retrieval}, pages 65--74, 2019.

\bibitem{zhang2016video}
Ke Zhang, Wei-Lun Chao, Fei Sha, and Kristen Grauman.
\newblock Video summarization with long short-term memory.
\newblock In {\em Proceedings of the European Conference on Computer Vision},
  pages 766--782, 2016.

\bibitem{zhang2016context}
Shu Zhang, Yingying Zhu, and Amit~K Roy-Chowdhury.
\newblock Context-aware surveillance video summarization.
\newblock {\em IEEE Transactions on Image Processing}, 25(11):5469--5478, 2016.

\bibitem{zhang_query-conditioned_2018}
Yujia Zhang, Michael Kampffmeyer, Xiaodan Liang, Min Tan, and Eric~P Xing.
\newblock Query-conditioned three-player adversarial network for video
  summarization.
\newblock In {\em 29th British Machine Vision Conference}, 2018.

\bibitem{zhao2018hsa}
Bin Zhao, Xuelong Li, and Xiaoqiang Lu.
\newblock Hsa-rnn: Hierarchical structure-adaptive rnn for video summarization.
\newblock In {\em Proceedings of the IEEE Conference on Computer Vision and
  Pattern Recognition}, pages 7405--7414, 2018.

\bibitem{zhao2019property}
Bin Zhao, Xuelong Li, and Xiaoqiang Lu.
\newblock Property-constrained dual learning for video summarization.
\newblock {\em IEEE Transactions on Neural Networks and Learning Systems},
  31(10):3989--4000, 2019.

\bibitem{zhou2018deep}
Kaiyang Zhou, Yu Qiao, and Tao Xiang.
\newblock Deep reinforcement learning for unsupervised video summarization with
  diversity-representativeness reward.
\newblock In {\em Proceedings of the AAAI Conference on Artificial
  Intelligence}, volume~32, pages 7582--7589, 2018.

\bibitem{zhou2020unified}
Luowei Zhou, Hamid Palangi, Lei Zhang, Houdong Hu, Jason Corso, and Jianfeng
  Gao.
\newblock Unified vision-language pre-training for image captioning and vqa.
\newblock In {\em Proceedings of the AAAI Conference on Artificial
  Intelligence}, volume~34, pages 13041--13049, 2020.

\bibitem{zhu2020dsnet}
Wencheng Zhu, Jiwen Lu, Jiahao Li, and Jie Zhou.
\newblock Dsnet: A flexible detect-to-summarize network for video
  summarization.
\newblock {\em IEEE Transactions on Image Processing}, 30:948--962, 2020.

\end{thebibliography}
